\documentclass[sigconf]{acmart}

\AtBeginDocument{%
	\providecommand\BibTeX{{%
			\normalfont B\kern-0.5em{\scshape i\kern-0.25em b}\kern-0.8em\TeX}}}

\usepackage{hyperref}
\usepackage{url}
\usepackage{graphicx}
\usepackage[ruled,vlined]{algorithm2e}
\usepackage[utf8]{inputenc} 
\usepackage[T1]{fontenc}    
\usepackage{hyperref}       
\usepackage{url}            
\usepackage{booktabs}       
\usepackage{amsfonts}       
\usepackage{nicefrac}       
\usepackage{microtype}      
\usepackage{tabularx} 
\usepackage{adjustbox}
\usepackage{color, colortbl}
\usepackage[utf8]{inputenc} 
\usepackage[T1]{fontenc}    
\usepackage{hyperref}       
\usepackage{url}            
\usepackage{booktabs}       
\usepackage{amsfonts}       
\usepackage{nicefrac}       
\usepackage{microtype}      
\usepackage{graphicx}
\usepackage{natbib}
\usepackage{color}
\usepackage{subcaption}
\usepackage{multirow}
\newcolumntype{L}{>{\raggedright\arraybackslash}X}

\begin{document}
	
\title{Solving Price Per Unit Problem Around the World: Formulating Fact Extraction as Question Answering}


\author{Tarik Arici}
\email{aricit@amazon.com}
\affiliation{%
	\institution{Amazon.com Inc.}
	\city{New York}
	\state{NY}
	\country{USA}
}

\author{Kushal Kumar}
\email{kushlku@amazon.com}
\affiliation{%
	\institution{Amazon.com Inc.}
	\city{Bengaluru}
	\state{KA}
	\country{India}
}

\author{Hayreddin \c{C}eker}
\email{hayro@amazon.com}
\affiliation{%
	\institution{Amazon.com Inc.}
	\city{Seattle}
	\state{WA}
	\country{USA}
}

\author{Anoop S V K K Saladi}
\email{saladias@amazon.com}
\affiliation{%
	\institution{Amazon.com Inc.}
	\city{Bengaluru}
	\state{KA}
	\country{India}
}

\author{Ismail Tutar}
\email{ismailt@amazon.com}
\affiliation{%
	\institution{Amazon.com Inc.}
	\city{Seattle}
	\state{WA}
	\country{USA}
}

\begin{abstract}
	Price Per Unit (PPU) is an essential information for consumers shopping on e-commerce websites when comparing products. 
	Finding total quantity in a product is required for computing PPU, which is not always provided by the sellers. To predict total quantity, all relevant quantities given in a product’s attributes such as title, description and image need to be inferred correctly. 
	We formulate this problem as a question-answering (QA) task rather than named entity recognition (NER) task for fact extraction. In our QA approach, we first predict the unit of measure (UoM) type (\emph{e.g.}, volume, weight or count), that formulates the desired question (\emph{e.g.}, ``What is the total volume?'') and then use this question to find all the relevant answers. Our model architecture consists of two subnetworks for the two subtasks: a classifier to predict UoM type (or the question) and an extractor to extract the relevant quantities. We use a deep character-level CNN architecture for both subtasks, which enables (1) easy expansion to new stores with similar alphabets, (2) multi-span answering due to its span-image architecture and (3) easy deployment by keeping model-inference latency low. Our QA approach outperforms rule-based methods by $34.4\%$ in precision and also BERT-based fact extraction approach in all stores globally, with largest precision lift of $10.6\%$ in the US store.
\end{abstract}


\maketitle

\section{Introduction}
\label{sec:introduction}
\label{intro}

PPU enables consumers to compare same or substitutable products when they come in a variety of packet sizes. Especially in the case of consumable products, PPU can be an important factor in customer purchase decisions. PPU information leads to a better consumer experience, and customers are sensitive to this especially in online retailing. This sensitivity leads to competitive pricing across online retailers regardless of the location of the customer and even moved offline retailers towards competitive pricing \cite{uniformpricing}. Grocery products with PPU information had higher purchase rate compared to those without PPU information. Also, some U.S. states and some EU stores such as UK have regulations on PPU reporting \cite{nist}. To this end, we train models that can correct wrong PPU information supplied by sellers and also fill in missing PPU information.

\begin{table}[t]
	\caption{Below, quantity information relevant to PPU task is shown in red for a few example catalog items.}
	\label{tagging}
	\scalebox{0.8}{
	\begin{tabular}{cc}
		\toprule
		Product Title & Total Quantity \\
		\midrule
		Maxwell House Original Roast Ground Coffee K-Cup Pods, \\ Caffeinated, \textcolor{red}{24} ct - 8.3 oz Box &24\\
		\midrule
		Maxwell House Original Roast Medium Ground Coffee, \\ Caffeinated, \textcolor{red}{42.5} oz Canister (\textcolor{red}{2} Pack) &85 oz \\		
		\midrule
		Tansukh Panchkol Powder for Hyperacidity and \\Digestion, red 60 gm (Pack of 2), (total \textcolor{red}{120} gm) & 120 gm \\
		\midrule
		Niconi Advanced Hand Sanitizer with 8 Hour Germ \\ Protection Lemon - \textcolor{red}{200} ml (pack of 2), (100 ml each) &200 ml\\
		\midrule
		Nutratech Creatine Monohydrate Micronized - \textcolor{red}{200} g\\ (Blueberry Flavor), 5000 mg Amino powder, \textcolor{red}{100} g extra &300 gm\\	
		\bottomrule
	\end{tabular}}
\end{table}

Worldwide, sellers often do not provide information about the total quantity directly but specify it in free-text fields like product titles, descriptions or even images. Moreover, this information is usually unstructured which renders regex-based extraction unfeasible which is highlighted in Table \ref{tagging}. In the first product title, the relevant quantity is $24$ while $8.3$ is not relevant in calculating the final quantity. In the second title, which is also a ground coffee product, all quantities are relevant. Even when there is some structure in the title, computing total quantity can be a challenge. For instance, the third and fourth product titles in Table \ref{tagging} follow similar syntactic structure, yet the context, words like \textit{total} and \textit{each}, decides the selection of the most relevant quantity. In the last product title although all quantities are related to \emph{weight}, $5000$ mg is not a relevant quantity as it indicates the concentration of the amino acid. These examples show that complex patterns, which depend on seller conventions, need to be learned, moreover, high-level information such as Unit of Measure (UoM) type is needed to guide quantity extraction. For example in the first product title knowing that the UoM type is count will increase the model's confidence to select quantity $24$ and not $8.3$ due to its context.

To the best of our knowledge, Named Entity Recognition (NER) is a de facto formulation for fact extraction problem, where the task is to assign tags to words in a sentence that indicates the begin and end of the answer. Traditionally, these named entities represent either a person, location or an organization \cite{ner}. We may formulate our quantity extraction problem similarly, in which we can either have a single extraction class (\emph{i.e.}, relevant quantity) or multiple classes corresponding to the relevant quantities in each UoM type. However, NER solutions do not enable the model to couple start and end indices explicitly, and check for their compatibility during training. Constraints such as the end index has to be bigger than the start index, can not be embedded during training. Moreover, they are prone to small variations in the tokens, for e.g., \emph{"fluid ounce"} or \emph{"fl oz"}, which need to be explicitly tagged. Other span characteristics such as shorter answers are more likely to occur, can not be learned by the model, which can help eliminating the need for post-processing or regularization. Also, UoM type information, which is shown to be important for quantity extraction, cannot be efficiently fed to NER model other than learning different token representations for each UoM type. To overcome these limitations, we introduce a span-image architecture that works at a character-level and employ a QA approach to quantity extraction which conditions the extractor model with UoM type information.

QA models such as the ones used in Stanford Question Answering Dataset (SQuAD) competition find an answer span given a context paragraph and a question. A significant advance in answer span prediction is the BiDAF method proposed in \cite{bidaf}, which uses bi-directional LSTM on query-to-context and context-to-query sequences and applies a softmax normalization across the sequence dimension. Hence, a single start index and a single end index is predicted to compute a single answer over context given query. BERT is a language model that significantly improved over the BiDAF model by enabling learning from unlabeled text \cite{bert} \cite{spanbert}. BERT and its alikes still use a softmax over the sequence as their goal is to find a single answer span. This is optimal since an answer span requires only one span start and end, and likelihood of start and end locations can be maximized separately. However, our task dictates as many quantity values as reported in product attributes (most likely up to three: weight/volume, items per package, and number of packages). Since we have to extract as many ``answers" as needed, we couple start and end prediction outputs and predict spans. We achieve this by outputting a vector for each location in a two-dimensional grid (span-image), where each location $(i,j)$ corresponds to a possible quantity span from character location $i$ to $j$. All acceptable spans occur in the upper triangular part of the span-image.

\begin{figure}[t]
	\centering
	\includegraphics[scale=0.33]{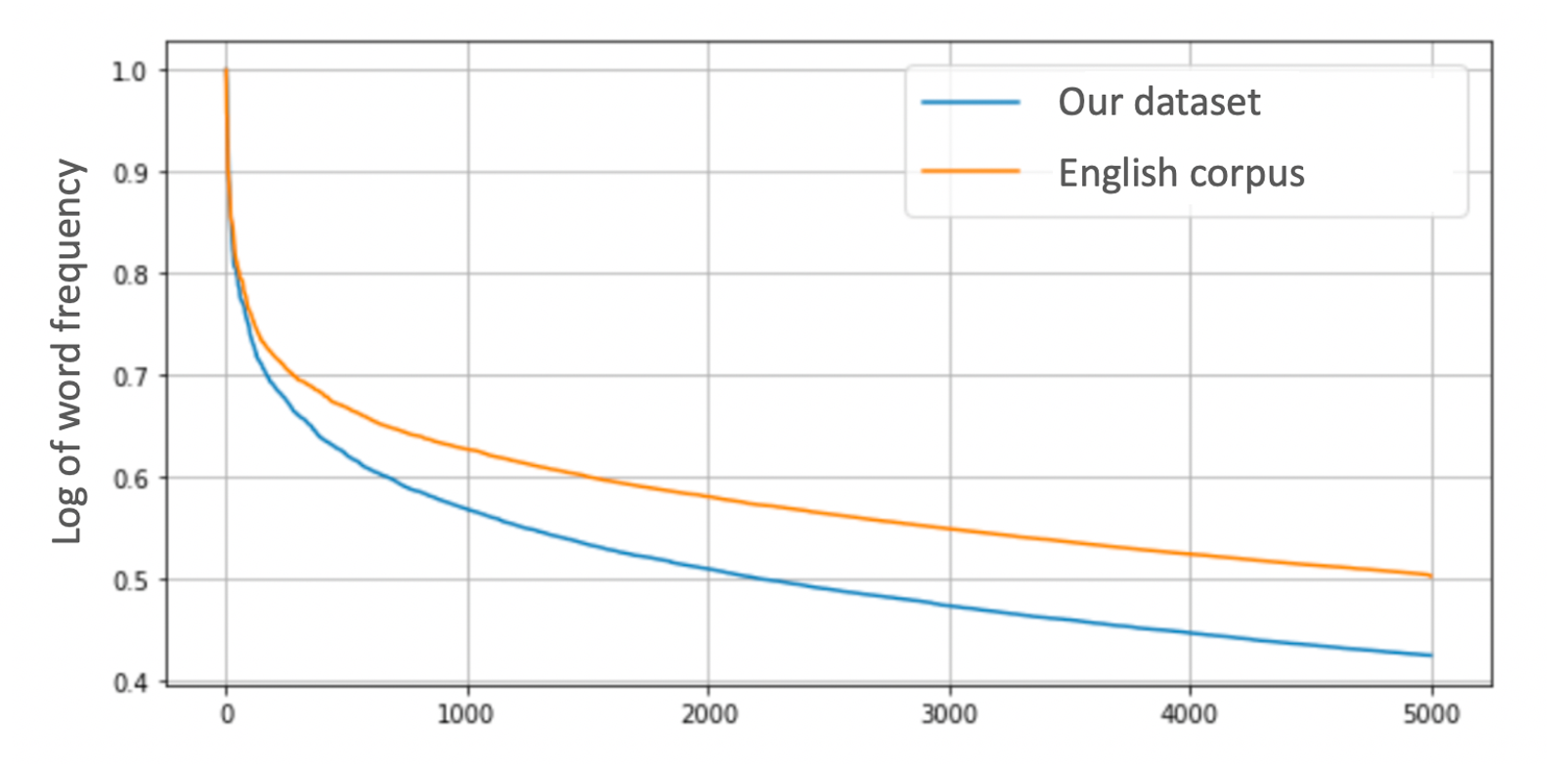}
	\caption{Word frequency plot for American English corpus and our dataset for top 5000 most common words. For an easier comparison, word frequency values are normalized in both corpuses so that log of word frequencies corresponding to most common words is one.} \label{word_hist}
\end{figure}

Our model uses available free-form textual attributes of a product, optical character recognition (OCR) text extracted from product images, and categorical features derived from product taxonomy. We designed a lightweight model that uses only character-level embeddings and convolutional layers, and is deep and large enough to learn semantic information to predict UoM type. Constructing a word-vocabulary and learning embeddings requires large number of parameters which is exacerbated by the heavy-tail distribution of words in our dataset (many rare words) as can be seen in Figure \ref{word_hist}. Unlike word vocabularies, a character vocabulary of size 128 applies to many stores with alphabets sharing characters. This enables sharing character-based models across various international stores and enjoy benefits of warm-start due to shared information such as brand names, measurements units, linguistic similarities, etc. Character-level convolutional networks have been successfully applied before to several text classification tasks and achieve state-of-the-art results for classification \cite{zhang2015character}. Our model has multiple use cases that can lead to enhanced quality of an e-commerce catalog by improving coverage and consistency of PPU related information in products display pages. These use cases include, but not limited to, defect rate reduction, backfilling missing values and real-time validation of PPU related information entered during registration of a new product (see Section \ref{sec:use_cases}).

Our contributions in this work are as follows:
\begin{itemize}
	\item We propose using a question-answering framework for extraction, where UoM classifier predicts a question that guides the quantity extractor. We employ a two-stage training approach. UoM classifier is trained in the first stage using more data available for this task, and quantity extractor is trained in the second stage by exploiting the predicted question as a latent variable. 
	\item We enable more than one possible answer spans in the input text by introducing a span-image architecture.
	\item We propose a character-based model deep enough to construct concepts and words from characters but light enough to satisfy our latency requirements and real-time use case.	
\end{itemize}

\section{Model Architecture}
\label{sec:architecture}
\begin{figure*}[t]
	\centering
	\includegraphics[scale=0.5]{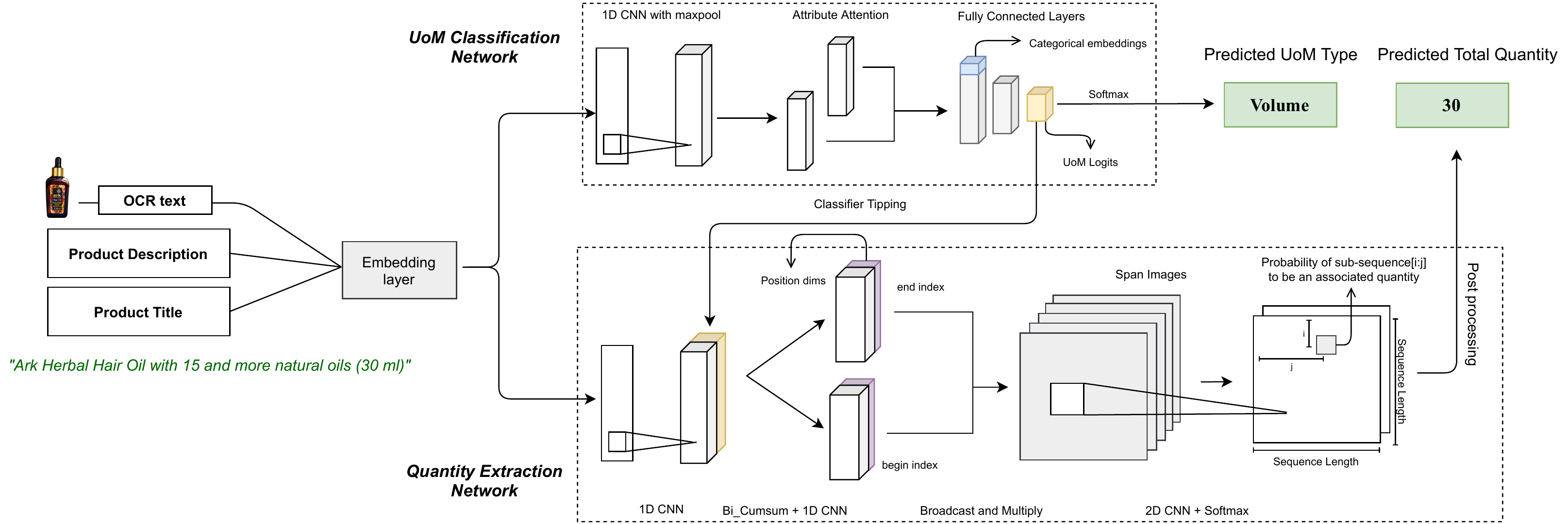}
	\caption{Our model architecture consists of two subnetworks: UoM classification and quantity extraction networks} \label{model}
\end{figure*}

Our neural network model consists of two subnetworks: UoM classification network and quantity extraction network as shown in Figure \ref{model}. Character embeddings are not shared between the two subnetworks since vocabulary size is small. Given text attributes $\{a^1, a^2, \cdots\}$, model input comprises of character sequences ${x}^i = \big\{{x}^i_1, {x}^i_2, \cdots, {x}^i_{n^i}\big\}$, where ${x}^i_j \in \mathbb{R}^k$ are $k$-dimensional character embedding corresponding to the $j^{th}$ character in the input sequence of length $n^i$ for attribute $a^i$. We also pass categorical embeddings as inputs for each category the product belongs to in the product taxonomy.

\subsection{UoM Classifier and Quantity Extractor}
UoM classifier consists of below stages (see Figure \ref{model}):
\begin{itemize}	
	\item Character embedding layer maps each character to a $k$ dimensional vector and the resulting vector sequence is fed into convolutional layers.
	\item Convolutional layers consist of multiple layers with filter sizes 3 and 5. We used \emph{maxpooling} on the activations. Output sequence vectors are batch normalized and dropout is employed.
	\item Attention module computes an attention vector from all input attributes. Each attribute-encoding vector is affine transformed to obtain attention keys, which are element-wise scaled and summed to find the scores for softmax weights. Weighted averaged attribute-encoding vectors constitute a product-description vector.
	\item Categorical embeddings vectors are created by embedding categorical indices into a high dimensional space ($1/\sqrt M$ dim, where $M$ is the number of categories). Every product in our catalog belongs to predefined categories. 
	\item Product-description vector and category-embedding vectors are concatenated and passed to classification layers to produce logits for UoM type. Softmax normalization is applied on the logits to predict the UoM type.
\end{itemize}
Quantity extraction (QE) model consists of below stages:
\begin{itemize}
	\item Character embedding layer same as above.
	\item 1D convolutional layers are applied to obtain an encoded sequence $\mathbf{y}$ without any strided pooling. Resultant sequence is batch normalized and dropout is used during training. No pooling is applied since sequence length needs to be kept.
	\item Each vector in $\mathbf{y}$ is concatenated with UoM softmax outputs, and fed into two different 1D convolutional layers to compute two vector sequences $\mathbf{s}$ and $\mathbf{e}$ of length $n$, with a shrunken depth $d$. This permits specialization for start and end index prediction. We also concatenate different positional dimensions to both these vector sequences to segregate them.
	\item $\mathbf{s}$ is tiled horizontally and $\mathbf{e}$ is tiled vertically to produce two tensors of size $n \times d$. These two tensors are multiplied element-wise to create a span-image of width and height equal to $n$ and depth of $d$. 2D convolutional filters are applied on the span-image to produce an image of size $n \times n$ and depth $2$. 
	\item Softmax normalization is applied on the depth dimension as opposed to the sequence dimension. Post-processing is done on the extracted quantities above a certain threshold to obtain the final quantity.
\end{itemize}

\subsection{Training and Inference}
\label{labeling}
Our model training is performed in two phases. In the first phase, UoM classifier is trained, in the second phase QE is trained while UoM classifier weights are frozen. This two-phase training strategy suits well with our model architecture. We use auditors to decide on UoM type and total quantity given a product. Our auditors specify a UoM type and total quantity value for each product audited. UoM type is a predefined class (\emph{i.e}, weight, volume, count). Hence, all available audits can be used for training the UoM classifier. However, total quantity is often a multiplication of other quantities such as number of items and/or packages, and item volume or weight. Since our auditors only provide final quantity value and do not explicitly tag parts-of-text within product attributes, we use some high precision heuristics (see Algorithm \ref{tagger}) to create the ground truth span required for quantity extraction model training. This approach of using heuristics to tag ground truth values does not work well for all samples, resulting in a small loss of audited examples. Our training dataset for learning the quantity extractor is about 15\% smaller than our UoM dataset for learning the classifier.

\begin{algorithm}
 \caption{Part-of-text tagging for quantity values in the attribute: We start searching for qualified candidates by evaluating combinations of upto 3 candidate quantities. This biases towards more factorial understanding of total quantity and favors learning features for number of packages, number of items, volume or weight. The qualified candidates are used for training quantity extractor model.}\label{tagger}
\textbf{Input}\\
$candidate\_span$: start \& end indices of candidate quantities\\
$total\_quantity$: total quantity determined by auditors\\
$uom\_type$: UoM type determined by auditors\\
\KwResult{$qualified\_spans$: trainable candidate spans}
\uIf{ $total\_quantity == 1$ and $uom\_type == count$}{
return $empty$ $list$}
\For{$k\in[3,2,1]$}{
\ForAll{$combination$ in ${nchoosek}(candidate\_span, k)$}{
\uIf{$\prod$(quantities $\in$ $combination$) $==$ $total\_quantity$}{
\uIf{no weight/volume quantity in $combination$ and $uom\_type == count$}{
return $combination$}
\uIf{only one weight/volume quantity in $combination$ and $uom\_type \neq count$}{
return $combination$}
}
}
}
\end{algorithm}

To increase our recall, we perform noising on our inputs by adding and deleting random gibberish words and tokens that are closely related to UoM classes. This leads to better generalization and introduces robustness to typos.

Our part-of-text tagging scheme for quantity extraction dataset is to find quantity candidates that can be processed to compute the final quantity using post-processing rules. At inference time, thresholding is applied on the span-image to obtain these candidate quantity span. Overlapping spans, if any, are handled by choosing the span with a bigger score. If no span has score higher than the threshold and UoM type is \emph{Count}, quantity is predicted to be one. In all other cases, the model refrains from making a prediction. 

We use rules to compute the final quantity from the obtained candidate quantities. Firstly, we identify the UoM type of each quantity span using cues from the immediate units of measure present in the text. If no unit of measure is found then UoM type is taken to be \emph{count}. We put all the obtained quantities in a stack and remove duplicate quantities of same UoM type. At every iteration for \emph{weight} or \emph{volume} UoM types we sum distinct quantities and remove quantities remaining in the stack which are duplicate to the sum obtained. For \emph{count} UoM type, we multiply the quantities from the stack and remove duplicates in the same way. Finally, we multiply total \emph{weight} or \emph{volume} quantity with the total \emph{count} quantity if the predicted UoM type is \emph{weight} or \emph{volume}, else output the total \emph{count} quantity.

\section{PPU Model Use Cases}
\label{sec:use_cases}
Our PPU model can be used primarily for the following three  tasks on an e-commerce catalog.

\subsection{Correction}
\label{defect}
This task involves fixing incorrect UoM and total quantity information provided by sellers. As this information may exist in one or more attributes, they may conflict with each other, e.g., having a different size in product title versus product image. We can use PPU models to predict UoM type and quantity for products at a particular cadence. If the prediction and associated attribute values in the catalog do not match, we can send them for manual correction. This can help remove defects, ensuring consistent information across attributes.

\subsection{Backfilling}
We can also use our model predictions for backfilling quantity information. For certain stores, we found correlations between the quality of catalog information and the popularity of the product. Specifically, the tail products tend to have inconsistent or missing attribute values along with distribution shifts on UoM types for same categories compared to head products. This leads to lower coverage at the time of backfilling. To address such problems, we experimented with \emph{active learning} technique by obtaining manual audits on a small set of tail products where the model confidence was low. It helped improve recall on tail products by $13\%$ with only $1\%$ drop on the head products in that store. 

\subsection{Validation}
\label{validation}
As new products are created in an e-commerce catalog everyday, the challenge of fixing incorrect information or backfilling missing values is ever lasting. Instead, we can ensure consistent attribute information during creation itself by using real-time model validation owing to low model latency. Quantity related attributes can be validated using our model predictions, and the merchant can be notified to recheck and correct inconsistent attribute values.

\section{Experiments}
\label{sec:experiments}

\begin{table}[t]
	\caption{Distribution of products by number of spans}
	\label{tab:span_length}
	\centering
	\begin{tabular}{cc}
		\toprule
		\# of spans & 	Percentage \\
		\midrule
		0 &	54.0\%\\
		1 &	34.5\%\\
		2 &	11.3\%\\
		3 & 0.2\% \\
		\bottomrule
	\end{tabular}
\end{table}

We evaluate our Quantity Extraction model based on its predicted total quantity and UoM type and not individual quantity spans. This is a stricter metric since a true prediction requires all relevant quantities in the input to be extracted correctly. We compare precision for quantity extraction task as we need to meet a high precision threshold for deployment, while for classification task, we compare F1 scores as both precision and recall are important.

\subsection{Dataset}

Our dataset was created within a 16-month time frame. We first used a rule-based model as a UoM classifier, and flagged items in the top $3$ product categories (we will refer to them as \emph{A}, \emph{B} and \emph{C} categories in this paper) when the prediction did not match the catalog values. Initially, we worked with internal audit teams to correct UoM and quantity values manually. After collecting about 40K examples in our dataset, we trained our deep learning-based UoM classifier, which had about 120K parameters. Designing a lightweight model allowed us to use deep learning technqiues early in our project. As we obtained more audits, our model size scaled proportional to our dataset sizes. Currently, our model includes 657K number of parameters. As the model performance improved, we started choosing candidates based on the correctness of quantity values as well as of the UoM type. Table \ref{tab:span_length} shows the distribution of products with respect to the number of spans. Large proportion of the products in our training dataset did not contain any span, which either meant that the relevant quantities are missing from the text or that the product is of type \emph{count} and the total quantity is $1$.\\

\begin{table*}\centering
	\caption{Performance gains for our PPU model over the rule-based model in US store\protect\footnotemark[3]}
	\label{tab:us_results}
	\begin{tabular}{*{20}{c}}\toprule
		\multicolumn{2}{c}{\multirow{2}{*}{Task}} & \phantom{abc} & {} & \multicolumn{4}{c}{PPU model ($\Delta$)} \\
		\cmidrule{5-8}
		& {} & {} && Volume & Weight & Count & Overall \\\midrule
		\multicolumn{2}{c}{\multirow{3}{*}{\textbf{UoM Classification}}} && $\Delta$P & {1.6} & {26.0} & {17.9} & {16.4} \\
		& && $\Delta$R & 64.0 & 56.9 & 90.0 & {76.3} \\
		& && $\Delta$F1 & \textbf{51.6} & \textbf{44.1} & \textbf{86.8} & \textbf{65.4} \\
		\midrule
		\multicolumn{2}{c}{\multirow{3}{*}{\textbf{Quantity Extraction}}} && $\Delta$P & \textbf{11.5} & \textbf{37.4} & \textbf{42.2} & \textbf{34.4}\\
		& && $\Delta$R & 19.1 & 0.2 & 27.7 & 19.1\\
		& && $\Delta$F1 & 22.4 & 8.0 & 40.6 & {26.1} \\
		\bottomrule
	\end{tabular}
\end{table*}

\label{in_data}
Similar to US catalog, IN catalog is also in English and shares the same vocabulary, yet there are several distribution differences when compared to US catalog. Fine-tuning US model even on a small training dataset can lift performance by increasing confidence scores on tokens including but not limited to unit words in the metric system for \emph{e.g. kilogram, millilitre,} etc. IN catalog is also rife with out-of-vocabulary (OOV) words for US model, which are borrowed directly from the regional language, for \emph{eg -} \textit{atta} which is Hindi for \emph{flour} and \textit{agarbatti} which is Hindi for incense sticks, that carry useful signals for UoM classification task. We also observed that longer text attributes such as \emph{product description} and \emph{bullet points} were seldom informative on PPU related information over shorter attributes like \emph{title}. Moreover, there were some distributional changes as well, like the distribution of UoM types across same product categories varied across the stores. Also, \emph{count} UoM type was more noisy where it was difficult to predict UoM type by using the product titles alone, and needed signals from product taxonomy. These differences needed to be addressed appropriately when testing US model and improving it further in IN store.

\begin{table*}[h]\centering
	\caption{Performance gains across stores over pre-trained Google BERT Base model fine-tuned first on catalog from English stores for both MLM and NSP tasks and then on UoM Classification task across all stores: EU-5, IN and US}
	\label{tab:bert}
	\scalebox{1}{
	\begin{tabular}{*{20}{c}}\toprule
		\multicolumn{2}{c}{\multirow{2}{*}{Store}} & \phantom{abc} & {} & \multicolumn{4}{c}{PPU Model Classifier ($\Delta$)} \\
		\cmidrule{5-8}
		& {} && {} & Volume & Weight & Count & Overall \\\midrule
		\multicolumn{2}{c}{\multirow{3}{*}{\textbf{EU-5}}} && $\Delta$P & -11.8 & -20.8 & -7.2 & -12.0 \\
		& && $\Delta$R & -9.0 & -7.8 & -3.9 & -5.4 \\
		& && $\Delta$F1 & {-10.2} & {-13.5} & {-5.6} & {-8.6} \\
		\midrule
		\multicolumn{2}{c}{\multirow{3}{*}{\textbf{IN}}} && $\Delta$P & 0.0 & 0.0 & 0.0 & 0.0 \\
		& && $\Delta$R & -8.0 & -4.4 & -3.9 & -5.0 \\
		& && $\Delta$F1 & {-4.4} & {-2.3} & {-2.1} & {-2.7}  \\
		\midrule
		\multicolumn{2}{c}{\multirow{3}{*}{\textbf{US}}} && $\Delta$P & 0.5 & -5.0 & 0.3 & -0.7 \\
		& && $\Delta$R & 5.8 & 12.3 & 3.6 & 6.8 \\
		& && $\Delta$F1 & \textbf{3.3} & \textbf{3.7} & \textbf{2.0} & \textbf{3.2} \\
		\bottomrule
	\end{tabular}}
\end{table*}

\begin{table*}\centering
	\caption{Problem formulation - Performance gains for Question Prediction and Answering approach using our PPU model over Fact Extraction approach using BERT across all stores: EU-5, IN and US}
	\label{tab:qa}
	\scalebox{1}{
	\begin{tabular}{*{20}{c}}\toprule
		\multicolumn{2}{c}{\multirow{2}{*}{Store}} & \phantom{abc} & {} & \multicolumn{4}{c}{Question Prediction and Answering ($\Delta$)} \\
		\cmidrule{5-8}
		& {} && {} & Volume & Weight & Count & Overall \\\midrule
		\multicolumn{2}{c}{\multirow{3}{*}{\textbf{EU-5}}} && $\Delta$P & \textbf{1.7} & \textbf{1.2} & \textbf{0.3} & \textbf{0.9} \\
		& && $\Delta$R & -27.3 & -20.2 & -15.3 & -19.6 \\
		& && $\Delta$F1 & {-30.7} & {-25.7} & {-17.2} & {-22.4} \\
		\midrule
		\multicolumn{2}{c}{\multirow{3}{*}{\textbf{IN}}} && $\Delta$P & \textbf{0.1} & \textbf{0.1} & -0.6 & \textbf{0.1} \\
		& && $\Delta$R & -14.1 & -7.4 & 11.3 & -0.5 \\
		& && $\Delta$F1 & {-8.7} & {-4.7} & {10.8} & {1.8}  \\
		\midrule
		\multicolumn{2}{c}{\multirow{3}{*}{\textbf{US}}} && $\Delta$P & -0.3 & -0.1 & \textbf{19.8} & \textbf{10.6} \\
		& && $\Delta$R & -8.3 & -4.8 & 8.5 & 1.6 \\
		& && $\Delta$F1 & {-7.6} & {-5.0} & {12.0} & {3.8} \\
		\bottomrule
	\end{tabular}}

\end{table*}

\subsection{Baselines to our PPU Model}
\subsubsection{Rule-based models}
The rule based model is the first baseline to PPU model, which comprises of regex rules for predicting UoM type and capturing all quantities present in the text. For predicting UoM type, the rule-based model relies on UoM specific keywords like \emph{ounce}, \emph{liquid}, \emph{pieces}, etc. A simple regex rule to capture weight quantities for instance, can be as follows: \emph{"[decimal number][space][weight unit]"}. Regex also catered to various composite patterns such as "$2 \times 200$ ml". We also applied guardrails to the per unit quantity value for each UoM type for better precision.

\subsubsection{Fine-tuned BERT models}
We fine-tuned BERT models separately for both the UoM classification and quantity extraction tasks as another baseline. We took pre-trained Google BERT base model (\emph{bert\_uncased\_L-12\_H-768\_A-12}) and trained it further on our catalog corpus for both MLM and NSP tasks for English stores like US and IN \cite{bert}. This model is then fine-tuned for the UoM classification task in the US store which has the largest share of training examples.

For quantity extraction task, we used BERT model to compare our approach with fact extraction formulation. We used BERT-base (\textit{bert-base-uncased}) model that is available from Transformers library which is trained on lower-cased English text with 12 layer, 768 hidden dimensions, 12 attention-heads and 110M parameters. We fine-tune it with PPU dataset as a fact extraction problem using a UoM agnostic question "What is the total quantity?". Current implementation of BERT does not support multi-span answer prediction. We modify the last linear layer and use two affine transformations (outer-product) to convert the separate begin and end vectors into a matrix where each pixel corresponds to a potential answer. The probability of each span can be computed by applying sigmoid function on each pixel in the output matrix. Using sigmoid makes no assumption on number of spans. This way, we can set a threshold to find out all plausible answers.

\subsection{Results}
We note that all results reported in this paper are in absolute terms.
\subsubsection{Comparison with rule-based models}
Table \ref{tab:us_results} shows performance comparison between rule based versus deep learning based PPU model in US store. Despite handling the most common quantity patterns in the rule-based model, rule-based model fails due to more complex patterns in the text and due to lack of semantic understanding of the product. We see that the deep learning approach significantly outperforms such rule-based models with F1 jump of over $65\%$ in UoM classification task and over $34\%$ in precision for quantity extraction task, where deep learning model crossed the set precision threshold for all UoM types but rule-based model did not. Also, the deep learning-based model has the potential to improve continuously as our corrections process yields more data as a byproduct while rule-based model has limited improvement potential (see Section \ref{defect}).

\footnotetext[3]{$\Delta$P - lift in micro-averaged precision, $\Delta$R - lift in micro-averaged recall}

\subsubsection{Comparison with BERT models}
Table \ref{tab:bert} compares performance in the UoM type classification task across all three stores between pre-trained BERT model and our UoM Classifier model. We compare F1 score for classification tasks, and as we can see in Table \ref{tab:bert}, our model outperforms in US store while performs somewhat comparably to BERT in IN store, although BERT model performs better in EU-5 store. Overall, despite any prior knowledge on English language or the e-commerce catalog and with much less parameters, our model comes reasonably close to BERT performance globally. This shows that our CNN architecture is deep enough to learn the semantic information to accurately predict UoM type.

As mentioned previously, the task of quantity extraction can also be viewed as a fact extraction problem, where the model directly predicts whether a certain quantity in the text attribute is relevant or not. We compare our Question Prediction and Answering approach on the lightweight PPU model against fact extraction approach using a bulkier BERT based model which was pre-trained for better English language understanding (see Table \ref{tab:qa}). We see that our model outperforms BERT in US store by $10.6\%$ and in EU-5 store by $0.9\%$ in precision while performing comparably in IN store, despite added advantage to the BERT model. Notably, our model crossed the set precision threshold for US store and on two out of three UoM types in both IN and EU-5 stores, while fact extraction using BERT crossed it only for \emph{weight} and \emph{volume} UoM types in IN and US stores. This reinstates that our approach of question prediction (in the form of UoM type) and answering is desirable since the UoM latent variables are critical in disambiguating among candidate quantity spans. It helps in achieving comparable performance to BERT with orders of magnitude smaller architecture which is easy to deploy.

\begin{table*}\centering
	\caption{Performance gains on 3 main Product Categories across different UoM types on quantity extraction task with respect to the overall performance in that store}
	\label{tab:prod_cat}
	\scalebox{0.97}{
	\begin{tabular}{*{20}{c}}\toprule
		\multicolumn{2}{c}{\multirow{2}{*}{Store}} & \phantom{a} & {} & \multicolumn{3}{c}{A} & \phantom{a} & \multicolumn{3}{c}{B} & \phantom{a} & \multicolumn{3}{c}{C} \\
		\cmidrule{5-7} \cmidrule{9-11} \cmidrule{13-15}
		& {} && {} & Volume & Weight & Count && Volume & Weight & Count && Volume & Weight & Count \\\midrule
		\multicolumn{2}{c}{\multirow{3}{*}{\textbf{EU-5}}} && $\Delta$P & 0.7& -15.1 & 7.9 && -1.3 & -3.3 & -12.9 && -7.0 & 0.1 & 8.1 \\
		& && $\Delta$R & -4.0 & -9.5 & 41.2 && 10.3 & 45.2 & 7.4 && -2.4& -0.8 &19.3 \\
		& && $\Delta$F1 & -5.9 & -15.8 & 41 && 13.2 & 47.4 & 7.1 && -3.8 & -1.2 & 16.6 \\
		\midrule
		\multicolumn{2}{c}{\multirow{3}{*}{\textbf{IN}}} && $\Delta$P & 0.1 & 0.0 & 0.9 && 0.2 & -0.1 & 0.7 && 0.2 & 0.0 & -0.3 \\
		& && $\Delta$R & -10.3 & -28.9 & -5.2 && -4.4 & 7.8 &-17.4 && -11.1 & -5.3 &-8.0\\
		& && $\Delta$F1 & -7.3 & -23.6 & -4.6 && -2.9 & 4.9 & -17.6 && -7.9 & -3.7 & -7.6 \\
		\midrule
		\multicolumn{2}{c}{\multirow{3}{*}{\textbf{US}}}
			 && $\Delta$P & 0.4 & 0.0 & 0.4  && 0.0 & -0.1 & 0.2  && 1.0 & -4.6 & 0.7 \\
		&	 && $\Delta$R & -4.7 & -0.5 & 12.0 	 && 15.7	 & 36.3 & -5.6 && -21.3 & 26.5 	& 9.8 \\
		&	 && $\Delta$F1 & -4.6	& -0.8 & 12.6 && 15.1 & 29.9 & -6.6	&& -24.7 & 21.5 & 10.5 \\
		\bottomrule
	\end{tabular}}
\end{table*}

\subsubsection{Product Categories}
Table \ref{tab:prod_cat} shows performance comparison across stores for the three most prominent product categories in our dataset - \emph{A}, \emph{B} and \emph{C}. Across all stores, product category \emph{A} has a particularly lower F1 score on \emph{weight} UoM type than the overall performance on that store. This is mainly because of incorrect UoM type classification since a lot of products in \emph{A} had a weight information in the product title but the audited UoM type was \emph{count}. For example a product title may look like - \textit{"Patanjali Saundarya Swarn Kanti Fairness Cream(1.75 Oz)"} where the model predicts its UoM type as \emph{weight} but the audited UoM type is \emph{count} and total quantity is $1$. For the same reason, recall on \emph{count} is higher in general for product category \emph{A} than the overall average. Also, we found that \emph{volume} and \emph{weight} UoM types have better precision than \emph{count} UoM type across categories. This is intuitive despite comparable classification performance across types as \emph{count} UoM type products usually have many more count related numbers (\emph{thread count, roll count, number of packs, etc.}) whose relevancy need to be accurately predicted. Due to the same reason, for EU-5 and IN stores, we are unable to reach the set precision bar for \emph{count} UoM type across all $3$ categories, unlike for \emph{weight} and \emph{volume} UoM types, using our PPU model.

\subsubsection{Feature Selection in IN}
As seen in Section \ref{in_data}, there are data distribution differences between US and IN stores. We found that a simple replacement of OOV words in IN store with US counterpart lifts US model's confidence by upto $10\%$ and even correct the predicted UoM type in some cases. Given our model is built on character-level features, it quickly adapts to the new set of words that are important for UoM classification task when fine-tuned (\textit{US fine-tuned (all text)} model), giving a substantial lift in recall by $26\%$ with respect to US baseline model. Inferring using only short text attributes such as \emph{title} with fine-tuned model (\textit{US fine-tuned (short text)} model) led to no notable drop in recall with improvements in \emph{weight} and \emph{volume} UoM types. Thus, the longer text attributes like \emph{product description} and \emph{bullet points} rarely contained extra information related to PPU over shorter attributes. 

Furthermore, we trained a model from scratch using all the text attributes (\textit{IN (all text)} model) and found that recall only increases by $16\%$ compared to US baseline model which is $10\%$ lower than the lift in US fine-tuned model. This also quantifies the impact of transfer-learning over training afresh, as the model transfers knowledge such as product type and brand names with similar UoM types from US store. We achieved best performance with a model trained from scratch using only short text attributes and categorical attributes (\textit{IN (short text \& categories)} model) with an overall recall lift of $36\%$ on held-out dataset with respect to US baseline. The role of categorical attributes is significant, where in its absence (\textit{IN (short text)} model) the recall lift is $10\%$ lower on the held-out set for quantity extraction task and $20\%$ lower on \emph{hard examples} for UoM classification task compared to the best model. We also trained a model from scratch using all text attributes as well as categorical attributes (\textit{IN (all text \& categories)} model) for completeness. This model performed reasonably well on the classification task on hard examples, yet failed on extraction task compared to the best model due to noisy signals from longer text attributes. See Table \ref{ablation} for results comparison across all the models.

\begin{table}[h]
	\caption{Performance gains on held-out dataset and \emph{hard examples} for IN across different model variants with respect to US baseline model. On both the held-out dataset and hard examples, we compare recall and F1 score at a fixed high precision point.}
	\label{ablation}
	\centering
	\scalebox{0.8}{
	\begin{tabular}{lccccc}
		\toprule
		\multirow{2}{*}{Model} & \multicolumn{2}{c}{Held-out Set (QE)} & \phantom{a} & \multicolumn{2}{c}{Hard Examples (UoM)} \\
	 \cmidrule{2-3} \cmidrule{5-6}
		& $\Delta$Recall & $\Delta$F1 && $\Delta$Recall & $\Delta$F1\footnotemark[4]\\
		\midrule
		US fine-tuned (all text) & 0.26 & 0.27 && 0.14 & 0.17\\
		US fine-tuned (short text) & 0.25 & 0.26 && 0.12 & 0.13\\
		IN (all text) & 0.16 & 0.17 && 0.14 & 0.17\\
		IN (all text \& categories) & 0.11 & 0.12 && 0.32 & 0.31\\
		IN (short text) & 0.26 & 0.26 && 0.16 & 0.18\\
		IN (short text \& categories) & \textbf{0.36} & \textbf{0.35} && \textbf{0.36} & \textbf{0.34}\\
		\bottomrule
	\end{tabular}}
\end{table}

\footnotetext[4]{Both $\Delta$Recall and $\Delta$F1 are micro-averaged metrics}

\subsection{Latency}
Model latency is a critical aspect for deployment, especially in the real-time validation use case (see Section \ref{validation}). The validation models are required to have low-latency (less than 50 milliseconds). Given our PPU model is designed to be light-weight and sufficiently deep, when compared to large language models like BERT, our model scales well with latency. On a machine with 2 CPU cores, mean latency for PPU model is $17\%$ better than BERT model and further improves as the number of CPU cores increases. Even with 16 CPU cores, we were unable to achieve less than $50$ milliseconds of mean latency for BERT model. Latency improves tremendously if we drop long text attributes like \emph{product description} \& \emph{bullet points} and use short text and categorical attributes, as in IN store (see Table \ref{latency_table}).

\begin{table}[h]
\caption{Latency scaling (in milliseconds) with respect to number of CPU cores. BERT and PPU Model rely on all the text attributes for inference, while PPU Model for IN use short text attributes and categorical features}
\label{latency_table}
\scalebox{0.88}{
\begin{tabular}{*7c}
	\toprule
	Number of CPU cores & \multicolumn{2}{c}{BERT} & \multicolumn{2}{c}{PPU model} & \multicolumn{2}{c}{PPU model - IN}\\
	\midrule
	{} & Mean & P90 & Mean & P90 & Mean & P90\\
	2 & 126 & 150 & 104 & 205 & 12 & 21\\
	4 & 73 & 89 & 56 & 105 & 8 & 13\\
	8 & 69 & 86 & 33 & 56 & 7 & 9\\
	16 & 56 & 66 & 21 & 32 & 6 & 7\\	
	\bottomrule
\end{tabular}}
\end{table}

\section{Conclusion}
\label{sec:conclusion}
We presented a lightweight deep learning model that can \emph{i}) perform semantic learning and  \emph{ii}) scale well with our dataset sizes and  \emph{iii}) be shared across different stores, thanks to its fully character based architecture. UoM for computing quantity depends on factors such as brand, product type, conventions in a store etc. This can only be learned from domain experts for each store through rigorous audit processes. As human labeled data is limited in size, large language models are not the best fit. A model fully scalable with dataset sizes is desired while low latency is a must for real-time use cases. Also, sharing models across stores is important as brands, product types, etc., have mostly same unit of measure. These restrictions coupled with the need to avoid tokenization errors makes fully character based architectures desirable, while span-image architecture allows multi-answering. Solving quantity extraction as a question prediction and answering task gives better performance over fact extraction formulation even with bulkier pre-trained language models like BERT.

\bibliographystyle{ACM-Reference-Format}
\bibliography{kdd_truefact_ppu}

\end{document}